# Landmark Tracking in Liver US images Using Cascade Convolutional Neural Networks with Long Short-Term Memory

Yupei Zhang[1], Xianjin Dai[1], Zhen Tian[2], Yang Lei[1], Jacob F. Wynne[1], Pretesh Patel[1], Yue Chen[3], Tian Liu[1] and Xiaofeng Yang[1,3,*]

[1]Department of Radiation Oncology and Winship Cancer Institute, Emory University, Atlanta, GA 30322

[2]Department of Radiation & Cellular Oncology, University of Chicago, Chicago, IL 60637

[3]The Wallace H. Coulter Department of Biomedical Engineering, Georgia Institute of Technology and Emory University School of Medicine, Atlanta, GA 30322

**Corresponding author:**

Xiaofeng Yang, PhD

Department of Radiation Oncology

Emory University School of Medicine

1365 Clifton Road NE

Atlanta, GA 30322

Tel: (404)-778-8622

Fax: (404)-778-4139

E-mail: xiaofeng.yang@emory.edu

**Abstract:** Accurate tracking of anatomic landmarks is critical for motion management in liver radiation therapy. Ultrasound (US) is a safe, low-cost technology that is broadly available and offer real-time imaging capability. This study proposed a deep learning-based tracking method for the US image-guided radiation therapy. The proposed cascade deep learning model is composed of an attention network, a mask region-based convolutional neural network (mask R-CNN), and a long short-term memory (LSTM) network. The attention network learns a mapping from an US image to a suspected area of landmark motion in order to reduce the search region. The mask R-CNN then produces multiple region-of-interest (ROI) proposals in the reduced region and identifies the proposed landmark via three network heads: bounding box regression, proposal classification, and landmark segmentation. The LSTM network models the temporal relationship among the successive image frames for bounding box regression and proposal classification. To consolidate the final proposal, a selection method is designed according to the similarities between sequential frames. The proposed method was tested on the liver US tracking datasets used in the Medical Image Computing and Computer Assisted Interventions (MICCAI) 2015 challenges, where the landmarks were annotated by three experienced observers to obtain their mean positions. Five-fold cross validation on the 24 given US sequences with ground truths shows that the mean tracking error for all landmarks is 0.65 ± 0.56 mm, and the errors of all landmarks are within 2 mm. We further tested the proposed model on 69 landmarks from the testing dataset that have the similar image pattern with the training pattern, resulting in a mean tracking error of 0.94 ± 0.83 mm. The proposed deep-learning model was implemented on a GPU, tracking 47 to 81 frames per second. Our experimental results have demonstrated the feasibility and accuracy of our proposed method in tracking liver anatomic landmarks using US images, providing a potential solution for real-time liver tracking for active motion management during radiation therapy.

## 1. Introduction

Accurate delivery of radiation dose to the intended treatment target is critical to the safety and efficacy of radiation therapy, especially in body sites where physiologic motion (respiration, sneezing etc.) may cause significant short-term variability in anatomic position (Lei *et al.*, 2020). Any inaccuracy may lead to insufficient does to the tumor target, geographic miss, or overdose of surrounding normal tissues. Many treatment protocols have attempted to reduce the impact of respiratory motion using breath-hold techniques. However, breath-hold substantially prolongs treatment time and may not be well-tolerated by all patients (Ha *et al.*, 2018). Real-time motion tracking enables advanced motion management during treatment delivery to improve treatment safety and efficacy while benefitting more patients.

Two-dimensional x-ray imaging is commonly used to assist motion tracing, but it often requires implanting several fiducial markers into the moving organs to facilitate tracking (Iwata *et al.*, 2017). Fiducial placement is invasive and carries with it a risk of side effects similar to those associated with other same-day operating room procedures. Recently, a few non-invasive tracking methods have been developed to track anatomic landmarks within moving organs with promising results (Huang *et al.*, 2019b; Ozkan *et al.*, 2017). Because it is non-invasive and low-cost while providing high soft tissue contrast in real time without additional radiation dose, US imaging is an excellent candidate for real-time imaging for motion tracking during radiation therapy (Zhang *et al.*, 2020a; van Sloun *et al.*, 2019). The automatic localization of landmarks potentially reduces the physician's cognition task and the manual error. However, US often suffers from low signal-to-noise ratio and imaging artifacts, making the motion tracking task on US images very challenging.

This landmark tracking problem is usually addressed by exploiting the relationship between the current image frame and the preceding frame in the US image sequence. Nouri and Rothberg trained a neural network by minimizing the Euclid distances between image patches containing the same landmark in an embedding subspace and then identified the target image patch with the shortest distance to the prior frame with a search window on the landmark (Nouri and Rothberg, 2015). Makhinya and Goksel extended the algorithm for superficial vein tracking using elliptical and template image sequences of the liver, followed by an optic-flow framework (Makhinya and Goksel, 2015). Hallack *et. al.* combined Log-Demons nonlinear registration to estimate motion with a moving-window tracking method to propagate motion around the region of interest (ROI) to subsequent frames (Hallack *et al.*, 2015). Kondo proposed two extensions to the kernelized correlation filter (KCF) using an adaptive window size selection and motion refinement with template matching (Kondo, 2015). Chen *et. al.* predicted the motion of anatomic targets in liver US sequences by a line regression-based ensemble of six machine learning models (Chen *et al.*, 2016). Ozkan *et. al.* proposed a supporter model to capture the coupling of motion between the image features and target so as to predict target position (Ozkan *et al.*, 2017). For 3D point-landmark tracking, Banerjee *et. al.* proposed a 4D US tracking method based on global and local rigid registration schemes (Banerjee *et al.*, 2015), while Royer *et. al.* combined visual motion estimation with a mechanical model of the target (Royer *et al.*, 2015). Williamson *et. al.* utilized a combination of template matching, dense optical flow and image intensity information for US target tracking in real-time (Williamson *et al.*, 2018). However, the above methods often suffer from abrupt motions due to sneezing, coughing, etc. To handle the tracking drift issues caused by abrupt motions, Teo *et al.* adopted a weighted optical flow algorithm to reduce the tracking drift of an uncontoured tumor (Teo *et al.*, 2019), while O'shea et al. used the $\alpha$-$\beta$ filter/similarity threshold for image-guided radiation therapy (Oshea *et al.*, 2016). In addition, Harris *et al* (Harris *et. al*, 2010) and Bell *et al* (Bell *et al.*, 2012) conducted the study of in vivo liver tracking using 4D ultrasound.

In recent years, deep learning-based methods have become the benchmark in a wide range of image processing tasks, such as image segmentation (Zhang *et al.*, 2020b; Lei *et al.*, 2021a), object detection (He *et al.*, 2017), image classification (Grimwood *et al.*, 2020) and image registration (Fu *et al.*, 2020). Gomariz *et. al.* proposed a fully convolutional Siamese network to learn the similarity between image patches related to the same landmark, where a temporal consistency model was built for regularization (Gomariz *et al.*, 2019). Huang *et. al.* used an attention-aware fully convolutional neural network to identify a suspect region and employed a convolutional LSTM to integrate temporal consistency in 3D US sequences (Huang *et al.*,

2019b). Then, Huang *et. al.* used a machine-learning based approach to generate subject-specific motion pattern and updated the template image using the learned motion pattern to reduce the search region (Huang *et al.*, 2019a). Liu *et. al.* proposed a one-shot deformable convolutional modal for enhancing the robustness to appearance variation in a meta-learning manner and combined this modal with a cascaded Siamese structure to enhance pixel-level tracking performance (Liu *et al.*, 2020). They also adopted an unsupervised training strategy to reduce the risk of overfitting on a limited sample of medical images. However, these methods are often lost in the similar image structures out of interest regions or missing the temporal features between frames. Dai *et. al.* developed a Markov-like network, which is implemented via generative adversarial networks, to extract features from sequential US frames and thereby estimate a set of deformation vector fields (DVFs) through the registration of the tracked frame and the untracked frames. Finally, they determined the positions of the landmarks in the untracked frames by shifting landmarks in the tracked frame according to the estimated DVFs (Dai *et al.*, 2021).

Mask R-CNN is a popular deep learning approach (He *et al.*, 2017), which has achieved state-of-the-art performance in object detection (Zhang *et al.*, 2020c; Lei *et al.*, 2021b). This study aims to attempt the mask R-CNN framework on the task of landmark tracking in 3D US imaging of the liver, where both advantages in these above methods are considered. The reason why we chose the popular mask R-CNN for motion tracking is that there are multiple landmarks with different image structures in an image frame. Besides, R-CNN has a capability of handling abrupt motions due to its less dependency of contexts. We set out to address two limitations of mask R-CNN when applied to this task: 1) incorrect proposals due to the structural similarity in local US image structures, as the cyan ellipses in Fig. 1; 2) limited single-frame scope that fails to exploit the temporal relationship between successive frames in a US image sequence. To address these limitations, an attention network is designed to focus the ROI of landmark, while a LSTM network is integrated to recognize landmark motion continuity. In the present study, we evaluate the proposed method on liver anatomic landmarks. Our methods and dataset are presented in detail in Section 2, experimental results in Section 3, followed by discussion and conclusions in Section 4.

## 2. Materials and Methods

### 2.A. Patient Dataset

In this study, we used data provided for the MICCAI 2015 Challenge on Liver Ultrasound Tracking (CLUST). The CLUST dataset is composed of 2D US image sequences acquired from 63 patients under a free-breathing protocol using 5 US scanners and 6 transducers. Based on the scanners used for acquisition, the US image sequences were divided into CIL, ETH, ICR, MED1 and MED2 groups, referring to the institutional image source (De Luca *et al.*, 2018). The duration of the sequence's ranges from 4 seconds to 10 minutes. The temporal resolution ranges from 6 Hz to 30 Hz, and the spatial resolution ranges from 0.27 mm × 0.27 mm to 0.77 mm × 0.77 mm. Anatomic landmarks were annotated on 10% - 13% of image frames per sequence, the number of landmarks range from one to five per sequence. These were manually annotated by three experienced observers; the resulting mean positions are used in this study. The 63 sequences were randomly divided into a training set of 24 sequences (40% of the dataset) and a test set of 39 sequences (60% of the dataset), yielding 53 landmarks for training and 85 landmarks for testing. For the test data, the annotation of landmarks in the first frame was provided to the trained network to track their positions in subsequent frames. Two example image frames are shown in Fig. 1. The regions indicated by cyan ellipse have a similar image pattern with the ground-truth landmark.

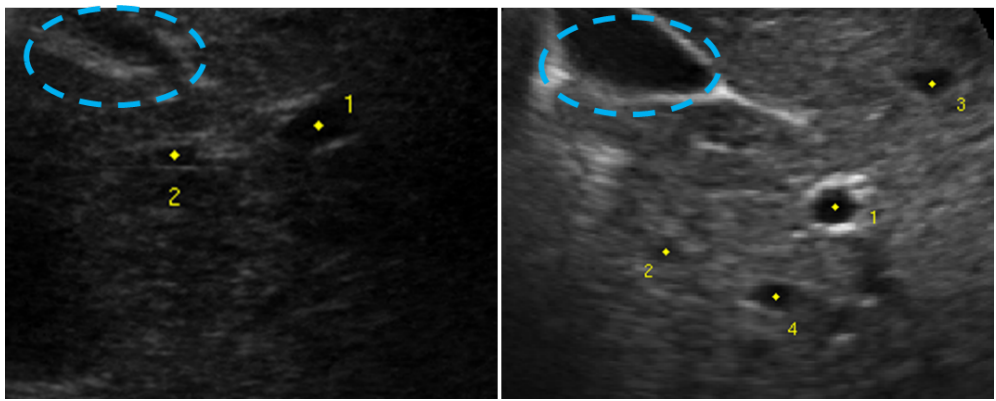

Fig. 1. Two example image frames of the CLUST dataset. The yellow points are the annotated landmarks. The cyan ellipse indicates the area of an ambiguous image structure that is seemed to be a landmark.

## 2.B. Attention Mask R-CNN with LSTM

The proposed deep model for landmark tracking is composed of three networks: an attention network, a mask R-CNN network and a LSTM network. The attention network learns a mapping from an US image to the attention area, where the landmark motion occurs, to reduce the search image region. The mask R-CNN network produces multiple ROI proposals for the landmarks in this region and identifies the landmark via three network heads: a bounding box regression, a proposal classification, and a mask segmentation. The LSTM network utilizes the US image sequence to model the temporal relationship between successive frames to assist bounding box regression and proposal classification. We integrated the three modules into an end-to-end deep learning architecture, as shown in Fig. 2.

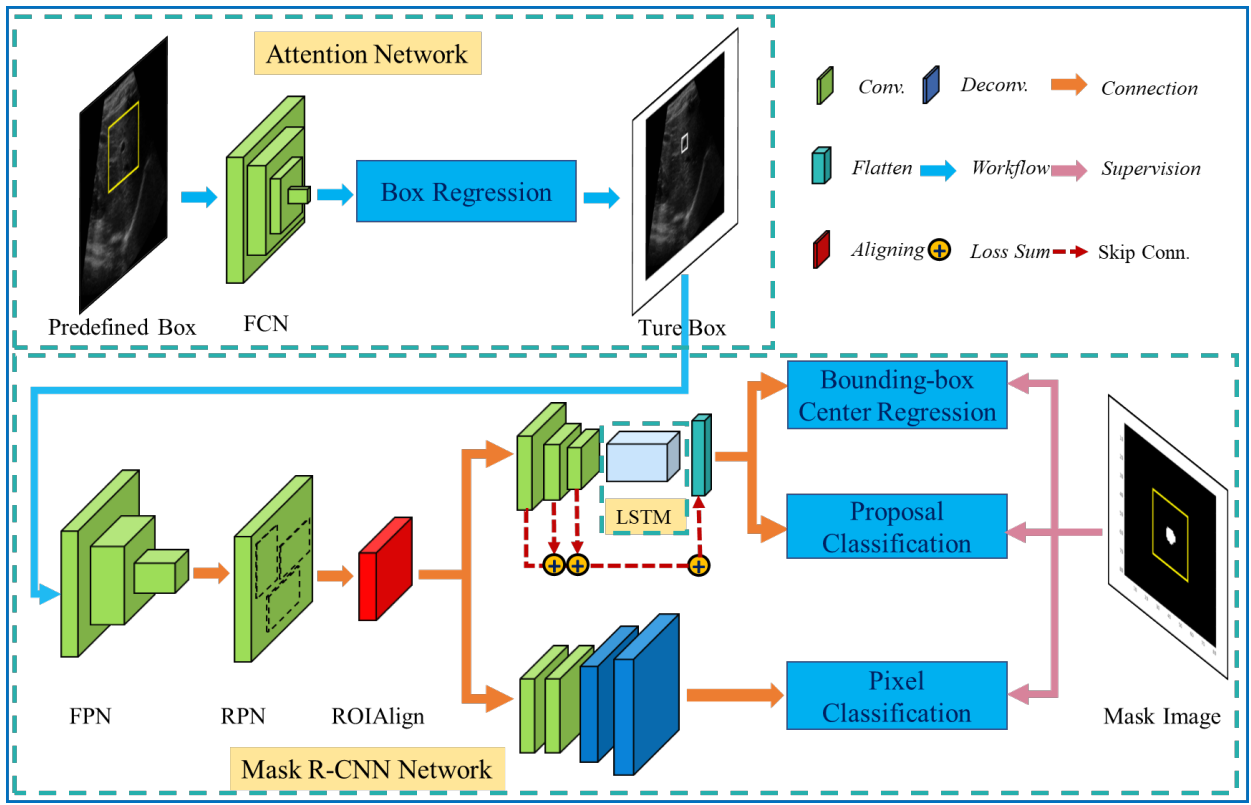

Fig. 2. The Proposed Deep Network Architecture, including Fully Convolutional Network (FCN), Feature Pyramid Network (FPN), Region Proposal Network (RPN) and Long Short-Term Memory (LSTM).

2.B.1. Attention network

The attention network aims to identify an attention ROI where the target landmark appears in order to reduce the search region considered by the subsequent mask R-CNN. Note that our attention network, whose aim is to learn a region box, is different from the work of Huang that adopted a pyramid attention network to learn the attention features (Huang *et.al* 2018). Since the landmark in the first frame is provided, we predefine a box that contains all its potential positions in the subsequent frames of the US sequence. A box of size 100 ×100 pixels centered on the landmark position in the first frame was selected as input for the attention network. To train this network, a box size of H × W pixels centered at the landmark of the target frame was designated as the ground truth. We extracted pairs of input and output boxes to learn the mapping that automatically reduces the search region. The number of pairs corresponds to all the landmarks provided in the training dataset. This box regression is made under the following objective function and the intersection over union:

$$L_{att} = min \sum_{i\in\{x,y,w,h\}}(t_i - o_i)^2 + \frac{1}{N}\|U - V\|_2^2 \tag{1}$$

where U and V are the input image patch and the output ground-truth patch, respectively; $N$ is the number of intersection pixels; $t_i$ and $o_i$ are the bounding box parameters of the target and the output, defined by:

$$t_x = (x - x_a)/w_a,\ t_y = (y - y_a)/h_a,\ t_w = log(w/w_a),\ t_h = log(h/h_a),$$

$$o_x = (x^* - x_a)/w_a,\ o_y = (y^* - y_a)/h_a,\ o_w = log(w^*/w_a),\ o_h = log(h^*/h_a),$$

where $x, y, w$ and $h$ are the coordinates of the box center, width and height, and $x$, $x_a$ and $x^*$ are the predicted box, the anchor box of a sliding window (Ren *et al.*, 2015) and the ground-truth box. Each anchor box, centered at the sliding window used in Mask R-CNN, is used as a reference to yield more region proposals. We adopted sliding-windows of different sizes to yield the proposal anchor boxes that have greater than 0.7 overlaps with the predefined box. After the attention network, the regions including irrelevant areas or ambiguous image areas are removed, as in Fig. 1, and greatly reducing the search cost within the target box.

2.B.2. Improved Mask R-CNN (IMask R-CNN)

Mask R-CNN is adopted to detect the bounding box centered at the target landmark (He et al., 2017), which is achieved by three training objectives: a bounding box regression $L_{box}$, a proposal classification $L_{cls}$ and a pixel classification $L_{mask}$ (Zhang et al., 2020c):

$$L = \omega_1 L_{cls} + \omega_2 L_{mask} + \omega_3 L_{box} \qquad (2)$$

where $\boldsymbol{\omega} = [\omega_1, \omega_2, \omega_3]$ was set to [0.2,0.2,0.6] in our studies to emphasize the bounding box regression.

While standard Mask R-CNN utilizes the softmax function for multi-object classification, object classification is formulated as a binary classification task by separating boxes containing or not containing a landmark. In addition to being better suited to a binary classification task, the large margin function proposed by Elsayed et. al. (Elsayed et al., 2018) has shown better performance compared with the softmax function. Therefore, we employed the large-margin loss function:

$$L_{cls} = \sum_{i=1}^{N} \sum_{l} max\left\{0, \gamma + \frac{f_o(x_i) - f_t(x_i)}{\varepsilon + \|\nabla_l f_o(x_i) - \nabla_l f_t(x_i)\|_2}\right\}, \qquad (3)$$

where $l$ is the $l$-th layer of the classification net, $\gamma$ indicates the decision boundary, and a small pertrubation $\varepsilon = 1e\text{-}6$ avoids the numerical instability. $f_o(\boldsymbol{x_i})$ and $f_t(\boldsymbol{x_i})$ are the scores classifying $\boldsymbol{x_i}$ into the class $o$ and its ground truth label $t$ respectively. Eq. (3) collects the margin losses at all layers for deep supervision (Zhang et al., 2020c).

Since mask images are composed of binary pixels, pixel classification is thus a binary problem. To improve pixel classification performance, we employed a large margin softmax loss initially proposed for binary pixel classification for needle localization in brachytherapy applications (Liu et al., 2016):

$$L_{mask} = \sum_{i=1}^{N} -log\left(\frac{e^{\|W_{y_i}\|_2 \|x_i\|_2 \varphi(\theta_{y_i})}}{e^{\|W_{y_i}\|_2 \|x_i\|_2 \varphi(\theta_{y_i})} + \sum_{j \neq y_i} e^{\|W_j\|_2 \|x_i\|_2 \cos(\theta_j)}}\right) + \lambda \|W\|_F^2, \qquad (4)$$

where $y_i$ is the ground truth label of $\boldsymbol{x_i}$, $j \in \{0,1\}$ indicates a non-landmark pixel or a landmark pixel respectively, $\mathbf{W}$ is the weights of the last fully connected layer, and $\varphi(\theta)$ is defined as (Liu et al., 2016):

$$\varphi(\theta) = (-1)^k \cos(m\theta) - 2k, \theta \in [\frac{k\pi}{m}, \frac{(k+1)\pi}{m}]$$

where $m$ is an integer that is closely related to the classification margin, and $k \in [0, m-1]$ is an integer.

For bounding box regression, we used center localization with a 20 pixels × 20 pixels bounding box (Zhang *et al.*, 2020c). The objective function is:

$$L_{box} = \sum_{i=1}^{N} \mathcal{L}(t_x^i - o_x^i) + \sum_{i=1}^{N} \mathcal{L}(t_y^i - o_y^i) \tag{5}$$

where $N$ is the number of samples in a mini-batch and $\mathcal{L}(u)$ is the robust loss function (He *et al.*, 2017) as:

$$\mathcal{L}(u) = \begin{cases} 0.5u^2, if\ |u| < 1 \\ |u| - 0.5, otherwise \end{cases}$$

2.B.3. LSTM Network

In an US image sequence, landmarks move continuously along successive frames, thus landmark position information obtained from one frame may improve localization in subsequent frames (Zhang *et al.*, 2020a; Bappy *et al.*, 2019). The power of LSTM networks for encoding context information and capturing temporal dependencies have been previously demonstrated (Greff *et al.*, 2016). The key components of the LSTM architecture providing this representational power are a memory cell that maintain its state over time and non-linear gating units which regulate information flow into and out of the cell. For each frame in an US sequence, each layer of the LSTM network computes $h_t$ at frame $t$ as,

$$i_t = \sigma(W_{ii}x_t + b_{ii} + W_{hi}h_{t-1} + b_{hi})$$
$$f_t = \sigma(W_{if}x_t + b_{if} + W_{hf}h_{t-1} + b_{hf})$$
$$g_t = tanh(W_{ig}x_t + b_{ig} + W_{hg}h_{t-1} + b_{hg})$$
$$o_t = \sigma(W_{io}x_t + b_{io} + W_{ho}h_{t-1} + b_{ho})$$
$$c_t = f_t \odot c_{t-1} + i_t \odot g_t$$
$$h_t = o_t \odot tanh(c_t)$$

where $h_t$, $c_t$, and $x_t$ are the hidden state, cell state and input at frame $t$ respectively, $h_{t-1}$ is the hidden state of the layer at frame $t$-1 or the initial hidden state, and $i_t$, $f_t$, $g_t$, $o_t$ are the input, forget, cell and output gates respectively. $\sigma$ is the sigmoid function and $\odot$ is the Hadamard product. The gates are used to transmit information from image frame to the next.

In this study, LSTMs were adopted to capture temporal features for each proposal, similar to Huang *et. al.* (Huang *et al.*, 2019b). During training, several proposals with the greatest overlap with the anchor box were considered to learn the gates in their corresponding cells. A similar LSTM network was used by Lei *et. al.* (Lei *et al.*, 2018). In the testing stage, proposals were fed into the corresponding LSTM to obtain the features, followed by fully connected networks for bounding box regression and object classification. Besides, we used LSTM for bounding regression and proposal classification but not for pixel classification to mitigate these bad effects from abrupt motion, such as cough and sneezing.

## 2.C. Similarity-based Localization Selection

The proposed approach (as shown in Fig. 2) often delivers multiple localizations with different scores for each landmark. However, the predicted localization with the highest score may correspond to confounding image structures that is similar to the target landmark. Since the movement of landmarks is smooth and continuous across successive frames, the landmark position at frame *t* can be inferred using the landmark position at frame *t*-1, based on the similarity between sequential frames. To consider both the score from IMask R-CNN and this prior, we improved our scoring schema by accounting for the distance of the landmark's predicted location to its location in the previous frame:

$$\boldsymbol{x}_t = arg \min_{\boldsymbol{x}_k}(\gamma S_k + (1-\gamma)\frac{1}{1+e^{\|x_k-x_{t-1}\|}}) \qquad (6)$$

where $\boldsymbol{x}_k$ is one localization with a score $S_k$ of all predictions at frame *t*, and $\boldsymbol{x}_{t-1}$ is the landmark at frame *t*-1. $\gamma$ is the trade-off parameter set to 0.5. Eq. (6) is composed of both mask R-CNN scores and the distances to the landmark at frame t-1. The combined score ranges from 0 to 1 and their contributions are controlled by $\gamma$, improving localization in our experiments.

**2.D. Model Training and Evaluation**

To train the proposed model, we set the learning rate to 1e-6, five continuous proposals for LSTMs and terminated the training at 1000 epochs where the decrease of the training error between two epochs is less than 1e-3. To evaluate our method on the training set of 24 US sequences, we used five-fold cross validation that divides the 24 sequences into five data subsets, *i.e.*, four subsets with five sequences per subset and one subset of four sequences. In each fold, we trained our model using four subsets and then tested on the remaining subset. To evaluate the model on the test data of 39 sequences, we trained the model using all 24 training sequences before submitting the results to CLUST organizer. The tracking error is computed on the predicted test landmarks using the Euclidean distance from the ground truth landmarks as follows:

$$err_i = \|\mathbf{t}_i - \mathbf{o}_i\|_2^2$$

where $\mathbf{t}_i$ is the ground truth position and $\mathbf{o}_i$ is the predicted position for a landmark in the *i*-th frame. The individual errors were calculated on all frames containing ground truth landmarks. Then, all errors for a landmark were used to compute the average error and standard deviation for each landmark. Finally, the average tracking error, standard deviation and the 95$^{th}$ error percentile were calculated for each patient. The computation cost was assessed to show the real-time imaging capability in US-guided radiation therapy.

## 3. Results

**3.A. Visualizations**

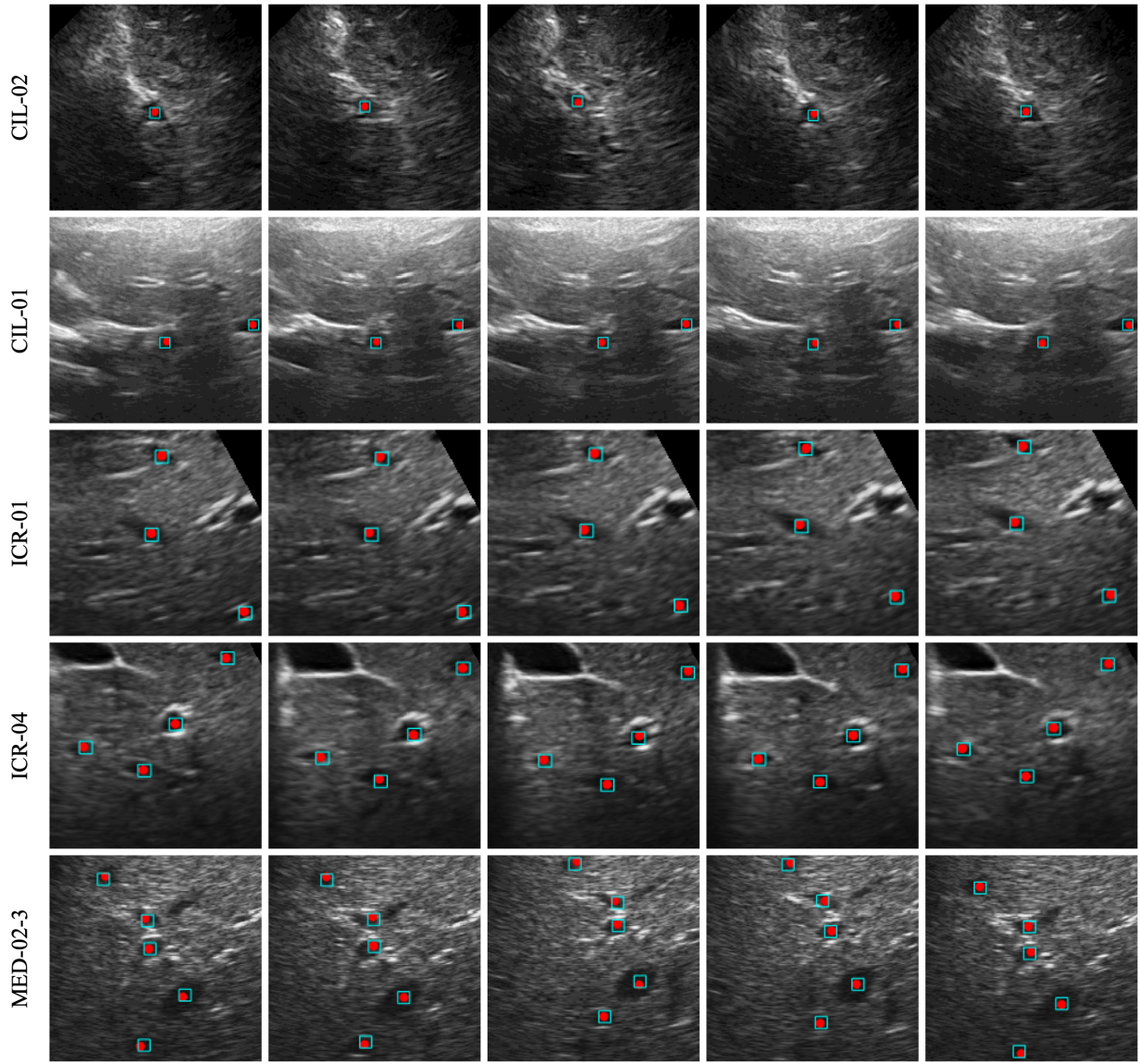

Fig. 3. The tracking results on five frames of five patients' US image sequence with one to five landmarks. The rows are the image sequences of five patients, where each set of patient sequences was captured with a different scanner. The patient identifiers are shown at left, while the five frames in a sequence are organized in columns. The red points are ground truth landmarks, and the cyan boxes are their predicted positions.

Fig. 3 shows tracking results on the US image sequences of five patients. Each was captured with a different scanner and contains one to five annotated landmarks. The tracking errors averaged over all frames are 0.79 ± 0.43 mm for patient CIL-02, 0.67 ± 0.58 mm for CIL-01, 0.71 ± 0.46 mm for ICR-01, 0.52 ± 0.31 mm for CIL-04, and 0.74 ± 1.14 mm for MED-02-3. As shown, the proposed method identified all landmarks on the five sequences regardless of the number of landmarks. A big mean error of 0.74 ± 1.14 mm is observed on the landmark within a shadow of patient MED-02-3, because the bounding box suffers from unsteadiness in the shadow. The tracking error is potentially caused by the inconsistency between landmark patterns and manual labels.

## 3.B. Quantitative Evaluation

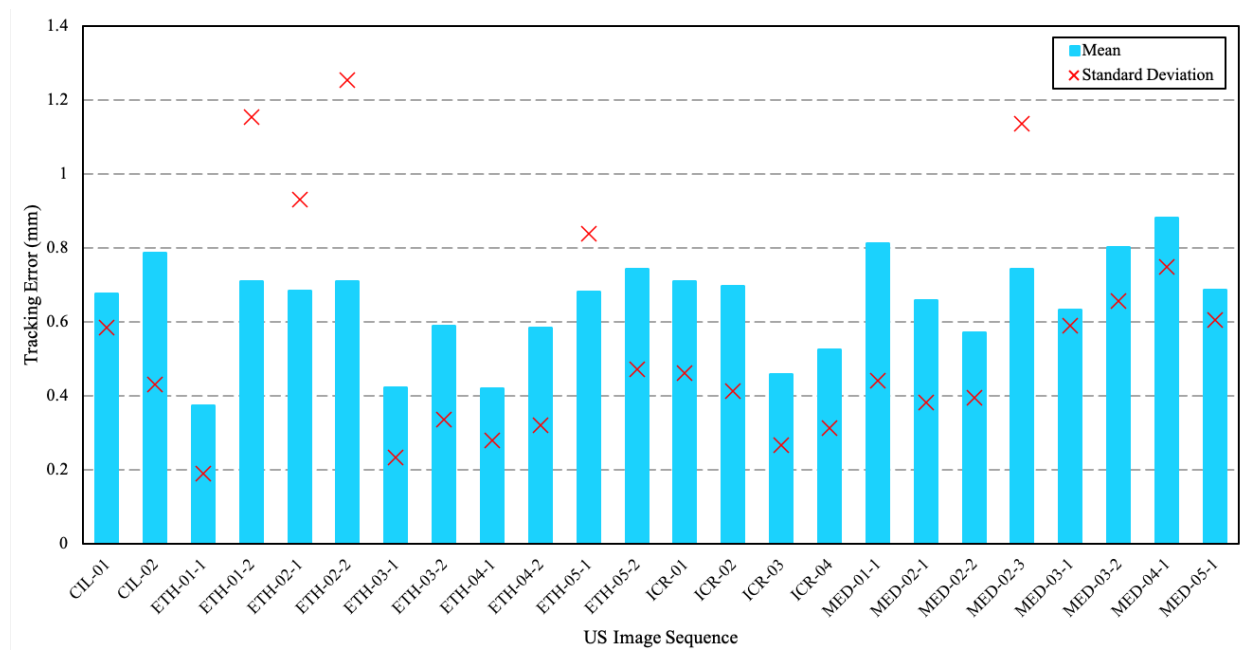

Fig. 4. Mean and standard deviations of tracking errors obtained on the 24 US sequences in training dataset.

Fig. 4 shows the mean error and standard deviation of tracking errors obtained on all image frames for each of the 24 training sequences, using five-fold cross validation. As shown, the mean tracking errors of all sequences are within 1 mm. For the sequences of ETH-01-1, ETH-02-1, ETH-02-2 and MED-02-3, large standard deviations are observed due to shadows surrounding these landmarks in the US images. Because a large shadow causes a freedom for the bounding box localization. The tracking errors on all landmarks is less than 2 mm, within acceptable clinical range (Keiper *et al.*, 2020; Huang *et al.*, 2019b). The minimum tracking error of 0.37 ± 0.19 mm was observed for ETH-01-1, while the maximum of 0.88 ± 0.75 mm was observed for MED-04-1.

Table I summarizes the evaluation results summarized on all 24 US sequences in terms of the mean, standard deviation, and 95[th] error percentile. The best performance was achieved in ICR sequences relative to other sources. On this training dataset, composite error of 0.65 ± 0.56 mm is reported for all sequence sources across all landmarks.

Table I. Summarized tracking errors per scanner source, including mean, standard deviation, and 95[th] error percentile. N denotes the number of landmarks in each scanner source. The last row summarizes the tracking errors averaged over all 53 landmarks in this study.

| Source | N | Mean (mm) | Std (mm) | 95% (mm) | AVE.MaxError(mm) |
|---|---|---|---|---|---|
| CIL | 3 | 0.73 | 0.51 | 1.66 | 1.82 |
| ETH | 16 | 0.59 | 0.60 | 1.52 | 1.64 |
| ICR | 12 | 0.59 | 0.36 | 1.39 | 1.52 |
| MED | 22 | 0.73 | 0.62 | 1.71 | 1.88 |
| ALL | 53 | 0.65 | 0.56 | 1.57 | 1.69 |

Table II summarizes the results on the training data with the proposed net components. To valid the effectiveness of the proposed attention, we removed the attention net from the proposed method which is called WAN (with attention) for short. We removed the LSTM net from the proposed method to valid its effectiveness, called WLSTM (with LSTM) for short. From the results, the proposed method is better than WAN and WLSTM. Both the attention mechanism and the LSTM have great effects on tracking results.

Table II. The comparisons between the three added components.

| Source | Mean (mm) | Std (mm) | 95% (mm) |
|---|---|---|---|
| WAN | 1.27 | 1.13 | 5.16 |
| WLSTM | 0.94 | 0.77 | 3.21 |
| Proposed | **0.65** | **0.56** | **1.57** |

## 3.C. Evaluation on Test Landmarks

We evaluated the proposed model on 69 test landmarks from the test dataset provided by this challenge organizer[1]. These test landmarks all have the similar image structure with the landmarks used for model training, where the image structure is shown in Fig. 3. Table III lists the evaluation results of these related methods. From the results, the proposed method achieves the better performance than other detection-based models, *i.e.*, Nouri's model (Nouri and Rothberg, 2015) and Gomariz's Model (Gomariz *et al.*, 2019). Fig. 5 shows the error distribution of the 69 test landmarks. There are 47 landmarks whose tracking errors are within 1 mm, and 15 landmarks whose tracking errors are in the range of [1 mm, 2 mm).

Table III. Evaluation comparison of the related models on the test dataset.

|                  | Mean (mm) | Std (mm) | 95% (mm) |
|------------------|-----------|----------|----------|
| Proposed         | **0.94**  | **0.83** | **2.43** |
| Nouri *et. al.*  | 3.35      | 5.21     | 14.19    |
| Gomariz *et. al.*| 1.34      | 2.57     | 2.95     |

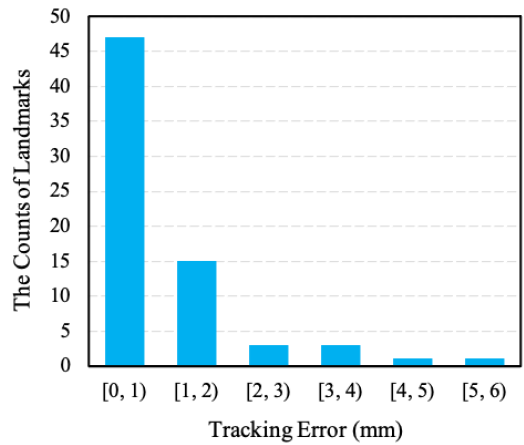

Fig. 5. The error distribution of the test evaluation using the proposed model.

---

[1] www.clust.ethz.ch/results.html

Fig. 6 shows the results of landmark localizations on an image sequence from the test data. The five frames show that the proposed method achieved a good performance on the left landmark, while obtained a weak localization on the right landmark. From observations, the left landmark has the same image structure with these landmarks used for model training, shown in Fig. 3. The right landmark has a greater shadow so that the localization is instable and inaccuracy.

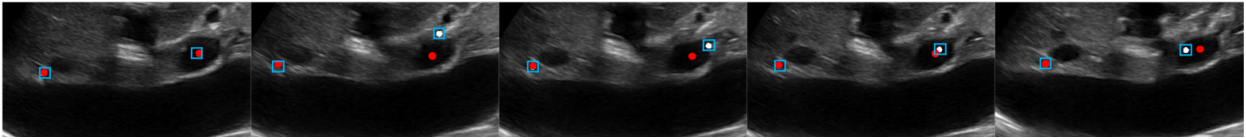

Fig. 6. An example of failed landmark tracking in the test data. Red points are manual ground truth, while cyan boxes are utilized to label the white points that is the predicted results.

### 4. Discussion and Conclusion

Landmark tracking is one of several common non-invasive motion tracking strategies, which can be used for advanced motion management during radiation therapy. This study proposes a deep learning-based approach composed of an attention network, an improved mask R-CNN, and an LSTM network to track landmarks on US image sequences. With the inclusion of an attention network, the search region can be greatly reduced. In the focused region, our method yields ROI proposals which are then conditioned by box regression, pixel classification and object identification. To improve the performance of Mask R-CNN for this binary classification, we implemented a large margin loss function with a deep supervision strategy for pixel classification and a large-margin softmax function for object identification. To exploit the temporal relationship between successive US frames, we also integrated an LSTM network into the Mask R-CNN.

Our experiments were conducted on the CLUST 2015 dataset, which was divided into a training set of 24 US image sequences and a test set of 39 sequences. Five-fold cross validation on the training dataset shows that our proposed method achieves an average error of $0.65 \pm 0.56$ mm with a 95% percentile tracking error of 1.57 mm. On the test set, evaluation results from the CLUST organizer shows that our method achieved $0.94 \pm 0.83$ mm on 69 test landmarks of the same with the training image structure. Besides, our proposed method could handle 47 to 81 frames per second, depending on the number of landmarks in the US image sequences. Comparing with other segmentation-based models, our method obtains the better performance on localization accuracy.

However, the proposed model has a limitation on tracking the landmarks with an unseen image structure, like other segmentation-based models (Gomariz *et al.*, 2019). Fig. 7 shows different landmark structures in the test dataset, where structure (a) is same with these structures in the training dataset shown in Fig. 3. Image structures (b, c) are out of the learned model. There are 9 images and 7 images on structure (b) and structure (c), where our model achieves $4.53 \pm 2.16$ mm and $8.81 \pm 5.37$ mm respectively. 13 images of these 16 images were given a more than 3 mm localization errors.

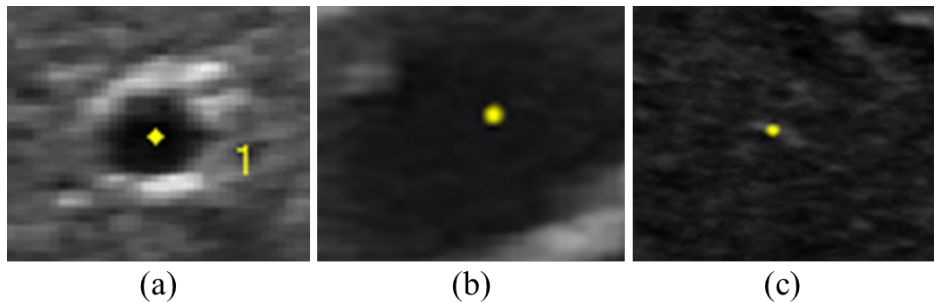

(a) (b) (c)

Fig. 7. The three landmark patterns observed in test sequences.

On the other hand, the registration-based method has reached a tracking result of 0.69 mm on the CLUST test dataset (Liu *et al.*, 2020). While our model is less accurate than the current best method, our model has already accepted performance on the localization accuracy and the tracking speed for a real-time clinical treatment system (Huang *et al.*, 2019b; Redmon *et al.*, 2016). Besides, our method is potential to be improved by training on more data and various landmark image structures. That is, the deep learning-based model is data hungry.

In summary, this study makes an attempt on deep learning-based method for real-time landmark tracking, resulting in a high accuracy with a small standard deviation on the landmarks known by the model. The proposed method allows whole image input so that tracking becomes a simple mapping operation. The performance is acceptable for real-time clinical applications. In future works, a U-Net could be integrated in the output head for pixel classification to exploit coarse and fine image feature scales. In addition to exploiting 3D features, U-Net might enhance the stability of localization results across image frames. One might also incorporate a registration network to further improve performance (Wang and Solomon, 2019). This study presents a potential method to address the problem of motion tracking for implementation in a real-time clinical system, suggesting advantages and disadvantages on the use of Mask R-CNN.

# ACKNOWLEDGMENTS

This research is supported in part by the National Institutes of Health under Award Number R01CA215718 and R01EB032680.